\documentclass[pmlr]{jmlr}% new name PMLR (Proceedings of Machine Learning)

\usepackage{longtable}% for long tables

\usepackage{bm}
\usepackage{times}
\usepackage{latexsym}
\usepackage{url}
\usepackage{xspace}
\usepackage{xcolor}
\usepackage{todonotes}
\usepackage{booktabs}

\usepackage{amsbsy}
\usepackage{adjustbox}
\usepackage[utf8]{inputenc} % allow utf-8 input
\usepackage[T1]{fontenc}    % use 8-bit T1 fonts
\usepackage{hyperref}       % hyperlinks
\usepackage{url}            % simple URL typesetting
\usepackage{booktabs}       % professional-quality tables
\usepackage{amsfonts}       % blackboard math symbols
\usepackage{nicefrac}       % compact symbols for 1/2, etc.
\usepackage{microtype}      % microtypography

\newcommand{\CXB}{\texttt{CheXpert++}\xspace}

\usepackage[load-configurations=version-1]{siunitx} % newer version
\makeatletter
\def\set@curr@file#1{\def\@curr@file{#1}} %temp workaround for 2019 latex release
\makeatother

\theorembodyfont{\upshape}
\theoremheaderfont{\scshape}
\theorempostheader{:}
\theoremsep{\newline}

\jmlrvolume{}
\jmlryear{2020}
\jmlrworkshop{Machine Learning for Healthcare (to appear)}

\title[\CXB]{\CXB: Approximating the CheXpert labeler for Speed, Differentiability, and Probabilistic Output}

\author{%
  \noindent \Name{Matthew B.A. McDermott} \\
  \Email{\texttt{mmd@mit.edu}} \\
  \addr CSAIL, MIT
  \AND
  \Name{Tzu Ming Harry Hsu} \\
  \Email{\texttt{stmharry@mit.edu}} \\
  \addr CSAIL, MIT
  \AND
  \Name{Wei-Hung Weng} \\
  \Email{\texttt{ckbjimmy@mit.edu}} \\
  \addr CSAIL, MIT
  \AND
  \Name{Marzyeh Ghassemi} \\
  \Email{\texttt{marzyeh@cs.toronto.edu}} \\
  \addr University of Toronto
  \AND
  \Name{Peter Szolovits} \\
  \Email{\texttt{psz@mit.edu}} \\
  \addr CSAIL, MIT
}

\date{}

\begin{document}
\maketitle

\begin{abstract}
It is often infeasible or impossible to obtain ground truth labels for medical data. To circumvent this, one may build rule-based or other expert-knowledge driven labelers to ingest data and yield silver labels absent any ground-truth training data. One popular such labeler is CheXpert~\citep{irvin_chexpert:_2019}, a labeler that produces diagnostic labels for chest X-ray radiology reports. CheXpert is very useful, but is relatively computationally slow, especially when integrated with end-to-end neural pipelines, is non-differentiable so can't be used in any applications that require gradients to flow through the labeler, and does not yield probabilistic outputs, which limits our ability to improve the quality of the silver labeler through techniques such as active learning.

% Old Result: 99.93\%

In this work, we solve all three of these problems with \CXB, a BERT-based, high-fidelity approximation to CheXpert. \CXB achieves 99.81\% parity with CheXpert, which means it can be reliably used as a drop-in replacement for CheXpert, all while being significantly faster, fully differentiable, and probabilistic in output. Error analysis of \CXB also demonstrates that \CXB has a tendency to actually correct errors in the CheXpert labels, with \CXB labels being more often preferred by a clinician over CheXpert labels (when they disagree) on all but one disease task. To further demonstrate the utility of these advantages in this model, we conduct a proof-of-concept active learning study, demonstrating we can improve accuracy on an expert labeled random subset of report sentences by approximately 8\% over raw, unaltered CheXpert by using one-iteration of active-learning inspired re-training. These findings suggest that simple techniques in co-learning and active learning can yield high-quality labelers under minimal, and controllable human labeling demands.
\end{abstract}

\section{Introduction}
In many non-mdecial settings, annotating even large datasets is relatively feasible via crowd-sourcing. In medicine, however, producing high-quality labels requires human expertise, and is thus much more expensive and impractical. To compensate for these difficulties, in several recent large radiology datasets~\citep{johnson_mimic-cxr:_2019,irvin_chexpert:_2019,bustos_padchest:_2019,wang_chestx-ray8:_2017}, researchers have relied on various rule-based, expert-defined labeling systems, most recently the CheXpert Labeler~\citep{irvin_chexpert:_2019}, which labels chest X-rays by running their corresponding free-text reports through a rule-based NLP model to predict either positive, negative, or uncertain mentions of 14 radiographic diagnostic categories. CheXpert has been used to define classification labels for two recent chest X-ray datasets \citep{johnson_mimic-cxr:_2019,irvin_chexpert:_2019}, and has been used to provide a metric for clinical accuracy in radiology report generation \citep{liu_clinically_2019}, all in the absence of any ground-truth labels for these tasks.

% there is a growing interest in applying machine learning to aid healthcare tasks ranging from clinical automation to decision support. However, these algorithms and systems typically rely on not only a huge number of input radiographs, but sets of patient- or exam-level golden clinical labels. Contrary to natural-image datasets which often resort to crowdsourcing for such labeling work, annotating medical images requires significant expertise, and is thus often prohibitively expensive and difficult. To get around this issue, researchers have designed and used rule-based natural language processing (NLP) tools to automatically derive clinical labels from free-text radiology reports which accompany and describe clinical radiographs, e.g., chest X-rays~\citep{peng_negbio:_2017,wang_chestx-ray8:_2017,irvin_chexpert:_2019}.

% One such system for chest X-ray annotation is the CheXpert labeler~\citep{irvin_chexpert:_2019} (which we will refer to as CheXpert).
%
%
CheXpert builds on top of NegBio~\citep{peng_negbio:_2017} which provides regular expression (regex) infrastructure for uncertainty and negation detection. 
% CheXpert also employs a set of complex decision rules for each of the sentences in the report before aggregating the results into final disease labels. 
As a result of being a regex-based system, CheXpert is not differentiable, and its outputs are binary labels, not continuous probabilities. CheXpert can also only be run on the CPU, rendering it relatively slow compared to GPU-optimized systems like neural network models.
% , thus losing the capability of being evaluated with discriminative measures such as area under curve (AUC).
% From an implementation perspective, these tools have several major caveats as well. In p as they often, in pursuit of ubiquity, rely on NLP tools written in Java. This prevents it from running efficiently and being able to integrate into current machine learning frameworks. Moreover, these tools fail to utilize GPUs which may limit their use cases to offline labeling.
%
CheXpert has been extremely valuable to the research community, but these three issues (lack of differentiability, probabilistic output, and runtime speed) induce significant hurdles for several exciting research directions. 

For example, \citeauthor{liu_clinically_2019} attempt to use CheXpert to enforce that automatically generated free-text radiology reports are clinically accurate by comparing their generated text to the ground truth in the CheXpert label space. However, as CheXpert is non-differentiable, they were forced to use a reinforcement learning policy gradient solution to optimize through the discrete labeler, a process which induced significant computational cost and, likely added additional instability to the results~\citep{liu_clinically_2019}. In a different direction, several notable researchers have also questioned the validity of CheXpert's labels~\citep{oakden-rayner_exploring_2017}, a finding we reproduce here. Were CheXpert to output probabilistic labels, we could attempt to improve these labels by employing active learning strategies leveraging the model's output uncertainty as a signal of which inputs most required human relabeling, or we could attempt to compensate for these label errors by using a small set of human labels to re-calibrate the model\footnote{Recalibration, being only an alignment between 1D inputs and 1D outputs, would require far fewer labels than would producing a set of labels for a new CheXpert} and then weighting our downstream use of these labels by their confidence scores. As it stands, we can do none of these things.

In this work, we solve these problem via \CXB, a Bidirectional Encoder Representations from Transformers (BERT; \citet{devlin_bert:_2018}) model trained to yield an extremely high-fidelity approximation to CheXpert. \CXB is differentiable, yields probabilistic outputs, and is fully GPU-ready.
\CXB is able to match CheXpert labels 99.81\% of the time on held-out radiology report sentences, so it can serve as a viable drop-in replacement for CheXpert, all while running
%in approximately 54\% of the time as 
approximately 1.8 times faster. Additionally, it is fully differentiable, and thus can be integrated natively into downstream neural pipelines. Finally, it yields probabilistic outputs, and therefore can be used with active learning systems. Unexpectedly, we also find that in the 0.07\% of cases where the two systems disagree, \CXB labels are actually more often preferred over CheXpert labels by a clinician, suggesting some inductive bias may be enabling \CXB to correct errors in its own training labels.

Building on that finding and to demonstrate the utility of this approximator, we additionally build a proof-of-concept active learning system and show that after only one epoch of re-training under solicited clinician annotations, \CXB is able to improve accuracy by 8\% in matching true clinician-defined gold standard labels over CheXpert. Note that in a concordant work, released while this paper was under review, \citeauthor{smit2020chexbert} has also found that mixed training a BERT based model on CheXpert labels and human labels can offer improvements to overall labeling performance. Whereas our active learning study is a proof-of-concept, their work uses a larger set of gold-standard annotations and should be regarded as a more full-fledged model aimed at improving labeler performance alone~\citep{smit2020chexbert}. 

In the rest of this work, we will first briefly outline the original CheXpert labeler and the construction, training, and performance of our approximation model \CXB, both as an approximation (e.g., in terms of fidelity to CheXpert) and in comparison to CheXpert via clinician evaluation. Next, we walk through in detail some projected use cases of this approximation model that would not be possible with the original CheXpert labeler, to better justify why we believe it is sensible to create such an approximation in the first place, focusing specifically on our proof-of-concept active learning system and the labeling improvements it imparts. Finally, we close with future work and concluding thoughts.

\subsection*{Generalizable Insights about Machine Learning in the Context of Healthcare}
Technically, this paper is (intentionally) quite simple. It uses existing models and algorithms, applying them in relatively standard ways. Our goal is not to demonstrate a new method or procedure for collecting labels or for modeling clinical text. Instead, our goal is two-fold:
\begin{enumerate}
    \item To provide as a resource a drop-in replacement for CheXpert that solves the 3 major hurdles identified above which inhibit novel research directions. Based on our own collective research history, we are confident this solution will prove useful to the community.
    \item To challenge the assumption that moving beyond silver labels is generally prohibitive, even if it requires additional annotations.
\end{enumerate}

This second point is our primary source of generalizable insight. Too often in machine learning for healthcare do we settle for sub-par labels before exploring procedures to intelligently and efficiently increase label quality, such as, for example, active learning. Evidence of this is the numerous works using the CheXpert labeler over the MIMIC-CXR dataset despite its known inadequacies.\footnote{To be clear, the authors of this work are not disparaging the CheXpert effort -- CheXpert has been extremely valuable to the community in enabling new kinds of research and facilitating new data release. Any purported flaws are opportunities to further benefit the field, not negative reflections on the original effort or execution.} In this work, with only roughly 3 hours of labeling time by a clinical coauthor and a single pass of fine-tuning, we were able to improve over the raw performance of CheXpert by approximately 8\% (averaged across all 14 CheXpert tasks) on a disjoint (also clinician-labeled) test set. The ease with which we obtain this gain is suggestive that these problems are not as insurmountable as they appear, and that machine learning for health, as a field, would be well served investing more in co-learning, model-approximation, and active-learning techniques to efficiently and rapidly improve label quality.

\section{From CheXpert to \CXB}

\paragraph{CheXpert Model Details}
% Though readers should refer to the CheXpert and NegBio papers for full details \citep{irvin_chexpert:_2019,peng_negbio:_2017}, we offer a brief sketch here of how these models work. 
CheXpert~\citep{irvin_chexpert:_2019} processes a span of text, and produces classification predictions across 14 different disease categories: 
\emph{Support Devices},
\emph{Airspace Opacity},
\emph{Atelectasis},
\emph{Cardiomegaly},
\emph{Consolidation},
\emph{Edema},
\emph{Enlarged Cardiomediastinum},
\emph{Fracture},
\emph{Lung Lesion},
\emph{No Finding},
\emph{Pleural Effusion},
\emph{Pleural Other},
\emph{Pneumonia},
and
\emph{Pneumothorax}.
For each disease category, CheXpert classifies the text as either containing 
\emph{no mention},
an \emph{uncertain mention} (indicating that the physician is uncertain),
a \emph{negative mention}, or
a \emph{positive mention}
of the disease category. 
% or \emph{no mention} (i.e., the top/ic is omitted).
% \stmharry{Maybe change the order to align with Figure 1: no, uncertain, neg, and pos}

CheXpert is an extension upon NegBio~\citep{peng_negbio:_2017}, a rule-based algorithm for detecting the 14 label categories used with the NIH Chest X-Ray 14 dataset~\citep{wang_chestx-ray8:_2017}.
% CheXpert differs from NegBio in 3 ways: First, CheXpert uses an expert curated list of terms for mention extraction (as opposed to ontologies such as MetaMap). Second, CheXpert uses more complex analyses to determine uncertainty and negation. Finally, CheXpert adds enhanced processing for double-negatives through pre- and post- negation analysis which improves over that of NegBio.
Being largely rule-based, CheXpert is non-differentiable and yields only predictions, not probabilities. Additionally, in practice, we have found it relatively slow and only suitable for offline labeling, even when run with many parallel processes. 

NegBio (in particular the labels assigned to the NIH Chest X-Ray 14 dataset) have been previously criticized for their inaccuracy \citep{oakden-rayner_exploring_2017}, so a valid question is whether these weaknesses extend to CheXpert.
\citet{irvin_chexpert:_2019} addressed these concerns by reporting comparisons to expert extracted labels for 1000 held-out radiology reports, finding strong performance for Mention F1 (0.948), and weaker but still strong performance for Negation F1 (0.899) and Uncertainty F1 (0.770). However, when tested on 687 reports from the (different) MIMIC-CXR dataset, \citet{johnson_mimic-cxr:_2019} found significant reductions in performance across all three tasks (0.874, 0.565, 0.470, respectively).

\paragraph{\CXB Model Details}
\CXB is a BERT based model, initialized from the pre-trained clinical BERT model~\citep{alsentzer_publicly_2019}, and followed by a multi-task prediction head. We trained \CXB at the per patient and report level over the MIMIC-CXR dataset~\citep{johnson_mimic-cxr:_2019}, using an 80/10 train/test data split (602,855 train sentences and 75,748/29,166 test sentences/not-in-train test sentences). Note two nuances to these data; first, that while we do split the data by unique report, due to the frequency of the use of templates in this modality, some individual sentences (which is the level at which we model) are shared across all splits. To account for this, we analyzed our results both on the overall test set and the unseen-sentences level (in which we simply removed the sentences that were also seen in training), finding nearly identical performance in both. Here, to be conservative, we report performances on the unseen-sentences test set, which had minutely worse numbers (overall parity on the full test set was 99.93\%, vs. 99.81\% on the unseen-sentences set) and imposes a slightly stronger generalization requirement at the expense of imposing some domain shift between the train and test sentences. Secondly, note that the first model trained yielded such strong performance that no hyperparameter tuning was necessary, so while we originally had an additional 10\% separated for validation, we did not use it at all in this work. The pre-trained BERT model was fine-tuned using 4 GPUs (GTX TITAN X) for 5 epochs using a batch size of 32 and an initial learning rate of $5\times10^{-5}$. Full training code and the pre-trained model are available.\footnote{\url{https://github.com/mmcdermott/chexpertplusplus}}

\paragraph{\CXB Model Results} \hfill

\noindent\textbf{Overall Performance}
\CXB approximates CheXpert with extremely high fidelity, matching CheXpert labels 99.81\% of the time on a held out set. This number is an average accuracy (match \%) over all 4 classes for all 14 labels. Each label individually also obtains over 99.7\% match accuracy, and all tasks save ``No Finding'' obtain over 99.9\% match accuracy. Especially when considering that CheXpert itself makes errors at rates far greater than this failure-to-match rate (as evaluated by~\citet{johnson_mimic-cxr:_2019,irvin_chexpert:_2019}), these performance numbers can justify \CXB as a direct replacement for CheXpert. Some readers may question why we don't just train \CXB directly on the underlying data backing CheXpert. The answer is simple: \emph{such data do not exist.} CheXpert is an expert-defined rule-based model, built and validated on separate, non-public datasets that we do not have access to. Using CheXpert as our training target is therefore very appropriate --- by matching at this rate we can confidently use \CXB as a drop in replacement, thereby solving our 3 main pain points, with no significant risk of disagreement all without requiring \emph{any} human-generated labels of any kind. Later in this work, we will discuss a proof-of-concept active learning system where we use an additional, human annotated set we construct to improve \CXB over CheXpert, but up to this point we rely on no human labels whatsoever.

\textbf{Per-task Performance Breakdown}
For more detailed performance numbers, Table~\ref{tab:per_task_error} lists the failure-to-match rate for all 14 labeling tasks, comparing majority class assignment performance (e.g., if one just predicted CheXpert's most frequent class, how often would this class match the CheXpert labels) to \CXB.
% In all cases, our error rates are lower by more than an order of magnitude.
We also break down performance across all labels for all tasks separately with per-task confusion matrices in the Appendix, Figure~\ref{fig:confusion_matrices}. We see some patterns emerge among our (very small) number of mis-classifications --- in particular, we very rarely mistake a ``No Mention'' label for a different label, whereas mis-classifications within the three other kinds of mentions are much more common.
%Additonally, we see that the only task on which we observe any visible deviation from purely diagnoal entries in the true label normalized view is support devices, and here we observe mis-classification between ``Uncertain Mention'' and ``Positive Mention.''

\begin{table}[t]
\centering
\caption{Failure-to-match rate (i.e., how often does \CXB disagree with CheXpert) for each task, for both a majority class classifier and \CXB. Smaller is better.}
\label{tab:per_task_error}
\small
\adjustbox{max width=\linewidth}{
% The table below is when we use the entire validation set, not just the distinct sentences.
% \begin{tabular}{lrr}
% \toprule
% Task                       & Majority Class & \CXB                   \\ \midrule
% No Finding                 & 33.55 \%       & $\boldsymbol{0.234\ \%}$ \\
% Enlarged Cardiomediastinum & 11.70 \%       & $\boldsymbol{0.082\ \%}$ \\
% Cardiomegaly               &  9.42 \%       & $\boldsymbol{0.085\ \%}$ \\
% Lung Lesion                &  0.96 \%       & $\boldsymbol{0.012\ \%}$ \\
% Airspace Opacity           &  8.11 \%       & $\boldsymbol{0.073\ \%}$ \\
% Edema                      &  6.13 \%       & $\boldsymbol{0.050\ \%}$ \\
% Consolidation              &  7.12 \%       & $\boldsymbol{0.071\ \%}$ \\
% Pneumonia                  &  3.33 \%       & $\boldsymbol{0.090\ \%}$ \\
% Atelectasis                &  5.63 \%       & $\boldsymbol{0.102\ \%}$ \\
% Pneumothorax               & 13.43 \%       & $\boldsymbol{0.059\ \%}$ \\
% Pleural Effusion           & 16.19 \%       & $\boldsymbol{0.096\ \%}$ \\
% Pleural Other              &  0.44 \%       & $\boldsymbol{0.011\ \%}$ \\
% Fracture                   &  1.31 \%       & $\boldsymbol{0.018\ \%}$ \\
% Support Devices            &  8.98 \%       & $\boldsymbol{0.061\ \%}$ \\
% \bottomrule
% \end{tabular}
% 
% This table is just on the distinct sentences.
\begin{tabular}{lrr}
\toprule
Task                       &  Majority Class       &  \CXB \\
\midrule
No Finding                 &           33.55 & $\boldsymbol{0.58}$ \\
Enlarged Cardiomediastinum &           11.70 & $\boldsymbol{0.20}$ \\
Cardiomegaly               &            9.42 & $\boldsymbol{0.21}$ \\
Lung Lesion                &            0.96 & $\boldsymbol{0.03}$ \\
Airspace Opacity           &            8.11 & $\boldsymbol{0.19}$ \\
Edema                      &            6.13 & $\boldsymbol{0.13}$ \\
Consolidation              &            7.12 & $\boldsymbol{0.19}$ \\
Pneumonia                  &            3.33 & $\boldsymbol{0.23}$ \\
Atelectasis                &            5.63 & $\boldsymbol{0.26}$ \\
Pneumothorax               &           13.43 & $\boldsymbol{0.14}$ \\
Pleural Effusion           &           16.19 & $\boldsymbol{0.25}$ \\
Pleural Other              &            0.44 & $\boldsymbol{0.03}$ \\
Fracture                   &            1.31 & $\boldsymbol{0.05}$ \\
Support Devices            &            8.98 & $\boldsymbol{0.15}$ \\
\bottomrule
\end{tabular}
}
\end{table}

\textbf{Error Analysis}
% \marzyeh{ I would be super brief about these results in the text because its a single person who rated things. This is not the meat of your contribution anyway, and you don't want a reviewer to get stuck on it.}
In addition to the analyses above, we also had a clinician colleague examine a random subset of up to approximately 46 discrepant examples per task (some tasks had fewer than 46 discrepancies total across the approximately 30,000 held out sentences, as noted in the table)\footnote{
Annotator was presented with a sentence, both labels (blinded), and indicated whether they preferred one label to the other, found both labels wrong, or were unsure (used 3 times of 528 total sentences annotated).
}. 
% \stmharry{Should we include micro-preference in the table? Besides, this 59\% might mislead the readers into thinking its 59 vs 41 for \CXB and CheXpert while it's only 29\% for CheXpert}
Perhaps surprisingly, we found that in a majority of cases (59\%), said clinician preferred \CXB labels over those of CheXpert, despite the fact that \CXB was trained to match CheXpert exactly. In 13\% of cases, both labels were deemed incorrect, and in only 28\% of cases was CheXpert preferred over \CXB.
% In roughly 13\% of cases, both labels were deemed incorrect, and in only the remaining 28\% of cases were the CheXpert labels deemed accurate. This likely reflects the fact that \CXB, being a continuous, neural model, has pre-built inductive biases which are less sensitive to mild phrasing alterations than the rule-based foundation of CheXpert proper.
A per-task breakdown of these results is in Table~\ref{tab:error_analysis}. We find that \CXB is preferred in all tasks save ``Lung Lesion'' (where CheXpert labels are preferred) and ``Pleural Other,'' where most often both are incorrect. 
%Note that these two tasks are the two tasks with the fewest discrepancies total (9, 8, respectively, out of the entire held out set), so these numbers may be more indicative of noise than a true task-dependent differential.
% Though these results are statistically underpowered (being driven by only one annotator) they may suggest that some sort of active learning or weak supervision approach may be particularly well suited here.

\NewDocumentCommand{\rot}{O{60} O{1em} m}{\makebox[#2][l]{\rotatebox{#1}{#3}}}%

\begin{table}[t]
    \centering
    \small
    \caption{Fraction of labels preferred by a clinician coauthor when CheXpert and \CXB disagree out of $N$ randomly chosen discrepant sentences. Note that $N$ is capped by the number of disagreements, which in several cases (denoted by a $*$) our annotator simply annotated all errors across the entire set. Larger means more preferred/better.}
    \adjustbox{max width=\linewidth}{
    \begin{tabular}{p{0.4\linewidth}rrrr}
        Task                       & N   &CheXpert      &\CXB      & Both Wrong     \\ \midrule %& Unsure
        No Finding                 & 44  & 25\%         & \bf 39\% & 36\%           \\ %& 0\% \\
        Enlarged Cardiomediastinum & 43  & 33\%         & \bf 56\% & 12\%           \\ %& 0\% \\
        Cardiomegaly               & 44  & 23\%         & \bf 61\% & 14\%           \\ %& 5\% \\
        Lung Lesion                & 9$^*$   &\bf 44\%      & 33\%     & 22\%           \\ %& 0\%\\
        Airspace Opacity           & 46  & 26\%         & \bf 61\% & 13\%	        \\ %& 0\% \\
        Edema                      & 38$^*$  & 29\%         & \bf 66\% & 5\%	        \\ %& 0\% \\
        Consolidation              & 46  & 30\%         & \bf 65\% & 4\%	        \\ %& 0\% \\
        Pneumonia                  & 45  & 22\%         & \bf 62\% & 16\%           \\ %& 0\% \\
        Atelectasis                & 46  & 24\%         & \bf 70\% & 7\%	        \\ %& 0\% \\
        Pneumothorax               & 41  & 22\%         & \bf 76\% & 2\%	        \\ %& 0\% \\
        Pleural Effusion           & 45  & 31\%         & \bf 58\% & 9\%	        \\ %& 0\% \\
        Pleural Other              & 8$^*$   & 38\%         & 13\%     & \bf 50\%       \\ %& 0\%\\
        Fracture                   & 14$^*$  & 21\%         & \bf 57\% & 21\%           \\ %& 0\% \\
        Support Devices            & 45  & 29\%         & \bf 62\% & 7\%	        \\
        \midrule
        Micro Average              &     & 27\%         & \bf 60\% & 12\%           \\
        \bottomrule
    \end{tabular}
    }
    \label{tab:error_analysis}
\end{table}

\textbf{Runtime}
\CXB is also faster than CheXpert. To label all MIMIC-CXR~\citep{johnson_mimic-cxr:_2019} sentences, CheXpert (using 32 processes) takes approximately 2.75 hours, whereas \CXB (using 1 GPU) takes only approximately 1.53 hours. We believe \CXB could also be optimized further, via neural network acceleration (e.g., \citet{lebedev_speeding-up_2018,cheng_model_2018}).

\section{Projected Use Cases of \CXB}
In addition to the raw speed improvement, which may justify its use directly, we envision several specific use-cases where having a differentiable, probabilistic model in place of CheXpert could be highly beneficial. We outline these at a high-level here, then explore a proof-of-concept system performing active learning using \CXB in Section~\ref{sec:active_learning}.
% \marzyeh{Maybe call this "Advantages of ChexBERT"?}
% \paragraph{Direct Labeling} Given its near perfect agreement with CheXpert, \CXB can be used to generate labels directly as an immediate stand in for CheXpert. Given its significantly faster speed, this would be valuable in its own right.

% \paragraph{Discriminative Measures of Performance} CheXpert (and the earlier NIH-specific variant) assess their performance using the F1 measure, which is a threshold-dependent performance measure. Given a true source of ground-truth labels, \CXB can be judged according to AUPRC or some other measure of discriminatve performance.

% \stmharry{+The tradeoff argument from previous notes}

\textbf{Active Learning/Weak Supervision} The automatic labeling tool used for Chest X-ray 14 \citep{wang_chestx-ray8:_2017} has been criticized for poor accuracy historically \citep{oakden-rayner_exploring_2017}. The CheXpert system, which builds on this system, may also suffer from similar issues. 
%Even if it does not, there is always room for improvement; however,
Given the cost of acquiring additional labels in this domain, techniques like active learning or weak supervision are extremely appealing. 
Prior works have found both sets of techniques helpful in processing hard-to-label data \citep{wang_cost-effective_2017,wang_deep_2018,halpern_electronic_2016, ratner2017snorkel}. Specifically within radiology report classification, \citet{nguyen_supervised_2014} study found dramatic annotation savings through active learning approaches. However, many active learning approaches require a notion of \emph{confidence} from the underlying classifier, which CheXpert alone cannot provide. As \CXB is fully probabilistic, it can be used natively with such active learning systems. We demonstrate an active learning proof-of-concept system with \CXB in Section~\ref{sec:active_learning}.
% \CXB can also integrate natively as a partially trained base learner in weak learning systems, or used in hybrid systems on both fronts.
%
% Our partial annotation results already suggest that through BERT's inductive biases, we're already yielding higher quality results than CheXpert in many cases, suggesting this approach may be particularly fruitful.

\textbf{Downstream Project Use}
% \stmharry{Upstream it provides gradient; downstream it provides continuous output}
Beyond its immediate utility in labeling, other works could also consider using the clinical labels provided by CheXpert or \CXB to enhance their radiograph processing systems in other ways. 
In our own prior work, for example, used CheXpert to assess the clinical accuracy of machine generated free-text reports for a radiograph. In order to get around the non-differentiable CheXpert labeler, we used a reinforcement learning (RL) policy gradient framework. As \CXB is fully differentiable, works like this could instead adopt \CXB in an end-to-end setting without any RL component, offering both a more direct objective function and a technically simpler solution~\citep{liu_clinically_2019}.
\citet{wang2018tienet} also included a text-based disease classification branch in their joint embedding learning network. With a pre-trained \CXB in place for such models, we save efforts in learning the text branch.
As \CXB is also much faster and can be run on a GPU, it also would yield a much more computationally performant integration than CheXpert does. This ability may help replace some of the more contrived methods of ensuring ``clinically accurate'' report generation that have been used, such as generating reports based on a training dataset composed exclusively of pre-filled templates~\citep{gale_producing_2018}.

\section{Active Learning Proof-of-Concept System}
\label{sec:active_learning}
In this section, we describe a proof-of-concept active learning system using \CXB, which ultimately improved ground-truth labeling performance by approximately 8\% (on average across all 14 tasks) as compared to CheXpert. While this system is not a full-fledged replacement labeler, it demonstrates that the active learning strategies enabled by \CXB can facilitate efficient, significant improvement in label quality. In this section, we will first detail how we structured our active learning experiments, then we will detail the results and implications of our findings.

\subsection{Active Learning Experiment Setup}
There are several key components to our active learning experiments. First, our new source of ground truth for evaluating the system against CheXpert; second, how we selected which examples to relabel for fine-tuning the \CXB system; and, finally, third, how we actually did the re-labeling and re-training.

\paragraph{Held-out Set Construction}
As stated previously, by default we have no gold-standard labels on these sentences; just the CheXpert defined silver labels. Thus, to enable evaluating our active learning system, we curated a set of 540 sentences for relabeling by 2 clinician colleagues in the following way.

First, for each CheXpert (task, label) pair (e.g., ``Edema, no mention'', or ``Fracture, positive mention''), we randomly selected ten sentences from the dataset which were assigned that label on that task by CheXpert. This amounted to 40 sentences for all tasks save ``No Finding'' (which only permits ``Positive Mention'' and ``Negative Mention'') which received 20 sentences. These 540 sentences served as our held out set, and each clinician annotated all of them, though each sentence was only annotated on its specific task (e.g., ``Edema''), to minimize annotation effort. The reason to do this tiered selection is to ensure that our held-out, golden-labeled subset spanned the breadth of conditions present in the reports as much as possible. Given performance numbers of CheXpert on this held-out set, however, we could use the overall propensity of these labels for their specific tasks to calculate the overall expected match rate, so we do not lose any information by doing this tiered selection process, and we gain significant diversity across tasks and expected labels.

When we compute CheXpert's raw match-rate against these labels, the results are not compelling; CheXpert is accurate only 69.8\%/71.6\% of the time on our held-out set according to each of our annotators individually. This echoes prior concerns about the validity of these automatic labelers~\citep{oakden-rayner_exploring_2017}. Note that the two annotators also disagreed non-trivially often (their agreement rate was 78.2\%), but they agree significantly higher with each other than either alone agrees with CheXpert.

\paragraph{Sample Selection}
In the interest of providing a simple proof-of-concept for the utility of \CXB's probabilistic outputs towards active-learning, we use a very straightforward selection process here. Namely, we selected, per task, the top 100 most uncertain samples under \CXB's probabilistic output within all reports in our validation set,\footnote{Note that, as sentences are duplicated within reports, this includes a subset of sentences that were in our original training set, but was fully deduplicated from the sentences in our annotated held out set} and asked our more senior annotator\footnote{We used just one annotator for the training set given the significantly increased burden} to (again in a per-task fashion) label all of these sentences, a process that took just over 3 hours. This resulted in 1086 new labeled sentences to use in training, each with only a subset of measured labels.\footnote{Note that this number is lower than the expected 1400 as many sentences are repeated across many radiology reports, and we eliminated these from our setup.} As we only used one annotator to do this re-training, we also evaluate with that annotator's held-out set to minimize sources of variance, though performance improvements (albeit much reduced) were observed on average in the other annotator as well despite the previously mentioned inter-annotator variance.

\paragraph{Retraining Procedure}
We fine-tuned \CXB for one additional epoch only over these 1086 sentences using 3 GPUs (note that using fewer GPUs would also work -- the system does not require significant GPU memory), only applying losses for labels that were actually set by our clinician annotator. This process took less than 30 seconds given how small the dataset was. Traditionally, in a full active learning pipeline, we would then solicit another round of "most-uncertain-samples" and repeat this procedure until performance reached sufficient level. Here, for our proof-of-concept, one epoch was enough to establish a significant performance gain.

\subsection{Active Learning Results}
\begin{figure}[!h]
    \centering
    \includegraphics[width=0.8\linewidth]{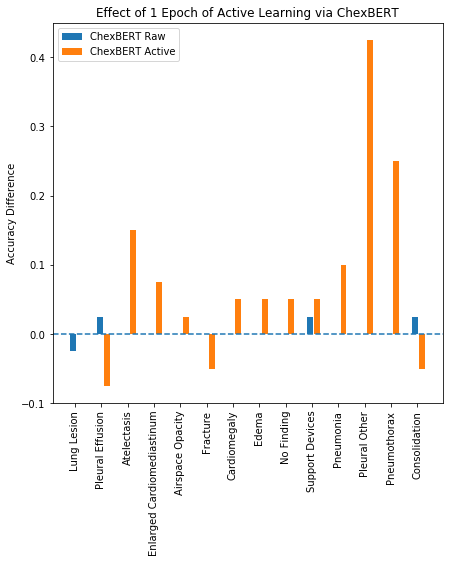}
    \caption{Accuracy on our clinician-labeled held out set of \CXB both with and without active learning minus accuracy of CheXpert on the same held-out set. Average accuracy of CheXpert is 71.6\%, average accuracy of \CXB raw is 72.0\%, average accuracy of \CXB post active learning is 79.1\%. Higher is better.}
    \label{fig:active_learning_gains}
\end{figure}

After this single epoch, performance improved on our new, gold-standard held-out set by approximately 8\% (averaged across all tasks) relative to both \CXB with no active-learning and CheXpert. Figure~\ref{fig:active_learning_gains} shows the difference in accuracy of both \CXB raw and \CXB post active learning vs. the raw CheXpert. We can see that even this single epoch of active-learning fine-tuning significantly improves performance on the majority of tasks, some quite significant margins.

These results show that with only 3 hours of annotation effort and 1 epoch of fine-tuning, we are able to reap significant label quality improvements via this simple approach.

\section{Discussion}
In this section, we first discuss several key takeaways from the work, then offer commentary on the next steps for this project. We then progress to concluding thoughts in the next section.

\subsection{Key Takeaways}
This work demonstrates both the utility of \CXB as a direct replacement for CheXpert in contexts where speed, differentiability, or probabilistic output matter, but also suggest that perhaps we need to place greater emphasis on strategies to intelligently and efficiently improve the quality of silver-labels in medicine. Many works have already used the CheXpert labels across different datasets, including~\citep{wang2018tienet,liu_clinically_2019,seyyed2020chexclusion}, among others, despite known issues relating to label quality. Here, we demonstrate first through the power of the apparent inductive biases of \CXB, as it improves on CheXpert's labels simply by training to match them, and second through our proof-of-concept active learning study, where we improve performance notably with minimal effort, that simple efforts could've appeased some of these label quality issues. This suggests we should place greater emphasis on similar strategies in general within machine learning for healthcare.

\subsection{Future Work}
\CXB has many advantages over CheXpert, but also has several notable opportunities for future improvements. 
Firstly, it has only been tested on one dataset, MIMIC-CXR. Testing on other datasets would be important for users to have full confidence in using \CXB on novel sources of radiology reports.
%
% Secondly, \CXB is currently significantly faster than CheXpert, but could likely be sped up much more with additional work on neural network optimization. Neural network compression and acceleration is an active area of research and could be explored here \citep{lebedev_speeding-up_2018,cheng_model_2018}.
%
Secondly, our model only operates at a per-sentence level---we have not investigated any methods for combining our per-sentence outputs into a global per-report set of labels, as CheXpert does. There are a number of strategies that could be explored in future work.
Finally, our partial manual annotation studies suggests that \CXB's ``errors'' are often corrections to CheXpert's labels, and our active-learning proof-of-concept strongly suggests that a full active-learning effort could likely yield higher quality labels with minimal additional annotator effort. This warrants significant additional analysis; using a full active learning analysis, with more annotators and a more in-depth error resolution procedure to raise the annotator agreement rate would be significant improvements and help transition our proof-of-concept into a full fledged system. Note that ~\citet{smit2020chexbert} perform a more involved annotation study in this vein in their (co-discovered) BERT based CheXpert improvement system, using 2 board certified radiologists and error resolution policies.

\section{Conclusion}

In this work, we presented \CXB, an extremely high fidelity neural approximation to the expert-labeling system CheXpert. We believe this work demonstrates that \CXB can serve as a faster, fully differentiable drop-in replacement for CheXpert, thereby enabling native integrations with downstream neural pipelines such as that of \citet{liu_clinically_2019}. Further, as it provides probabilistic output, it can be used with active learning algorithms to refine original label quality for classificaiton purposes.
% Lastly, \CXB is significantly faster than CheXpert.
These properties will render \CXB useful in a variety of settings as the community continues to make increasing use of the chest X-ray modality. We demonstrate this likely utility both by outlining several strategies of projects that could benefit from \CXB and via an active-learning study which improved expert-defined gold standard performance by an average of 8\% as compared to CheXpert across all 14 labeling tasks.

\section{Acknowledgments}
This work would not be possible without the aid of Dr. Catherine McDermott
who, in addition to the critical support of coauthor Dr. Wei-Hung Weng in this regard, helped us annotate various examples between \CXB and CheXpert. Additionally, this work is funded in part by National Institutes of Health: National Institutes of Mental Health grant P50-MH106933 as well as a Mitacs Globalink Research Award.

\bibliography{ChexBERT}

\clearpage
\appendix
\section{Further Experiments in CheXpert \CXB Parity}

\begin{figure*}[!ht]
    \centering
    \includegraphics[width=\linewidth]{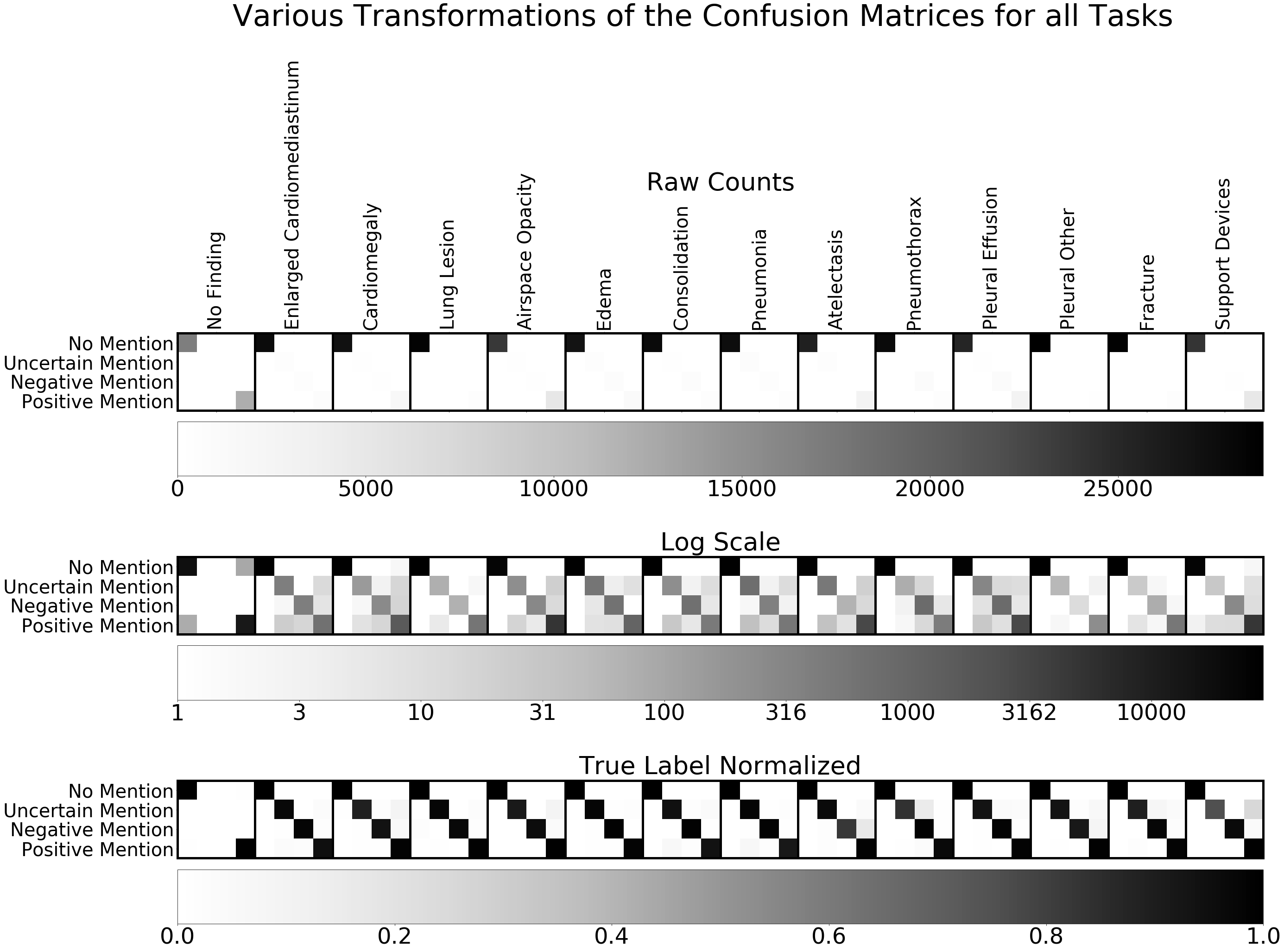}
    \caption{Confusion matrices of \CXB against CheXpert in raw counts, log-scale counts, and reference-label (i.e.,\ row) normalized ratio for all tasks. 
    Note that each of the labeling tasks and the associated confusion matrices are independent. The vertical axis corresponds to the reference (CheXpert) label while the horizontal axis in each matrix corresponds to predictions from \CXB.
    % Note that we have stacked all tasks in one row for each set of confusion matrices, but they are all independent prediction tasks. 
    Log scale is added to better illustrate the breakdown of the very small number of mis-classifications observed.}
    \label{fig:confusion_matrices}
\end{figure*}

% \section{Additional Active Learning Results}

% \begin{table}[!t]
%     \centering
%     \begin{tabular}{lrrrrrr} \toprule
%         Task & \begin{multicol} \\
%          & 
%     \bottomrule \end{tabular}
%     \caption{Caption}
%     \label{tab:my_label}
% \end{table}

\end{document}